\documentclass{article}

\newcommand{\PCignore}[1]{}

\newcommand{\squishlist}{
 \begin{list}{$\bullet$}
  { \setlength{\itemsep}{0pt}
     \setlength{\parsep}{3pt}
     \setlength{\topsep}{3pt}
     \setlength{\partopsep}{0pt}
     \setlength{\leftmargin}{1.5em}
     \setlength{\labelwidth}{1em}
     \setlength{\labelsep}{0.5em} } }

\newcommand{\squishlisttwo}{
 \begin{list}{$\bullet$}
  { \setlength{\itemsep}{0pt}
     \setlength{\parsep}{0pt}
    \setlength{\topsep}{0pt}
    \setlength{\partopsep}{0pt}
    \setlength{\leftmargin}{2em}
    \setlength{\labelwidth}{1.5em}
    \setlength{\labelsep}{0.5em} } }

\newcommand{\squishend}{
  \end{list}  }

\usepackage{microtype}
\usepackage{graphicx}
\usepackage{subfigure}
\usepackage{booktabs} 
\usepackage{amsmath}
\usepackage{mathtools}
\usepackage{graphicx}
\usepackage{comment}
\usepackage[table,xcdraw]{xcolor}
\usepackage{multirow}

\pdfoutput=1

\usepackage{hyperref}

\usepackage[accepted]{sysml2019}

\sysmltitlerunning{Conditional Neural Architecture Search}

\begin{document}

\twocolumn[
\sysmltitle{Conditional Neural Architecture Search}



\sysmlsetsymbol{equal}{*}

\begin{sysmlauthorlist}
\sysmlauthor{Sheng-Chun Kao}{to}
\sysmlauthor{Arun Ramamurthy}{goo}
\sysmlauthor{Reed Williams}{goo}
\sysmlauthor{Tushar Krishna}{to}

\end{sysmlauthorlist}

\sysmlaffiliation{to}{Georgia Institute of Technology, GA, USA}
\sysmlaffiliation{goo}{SIEMENS Corporate Technology, NJ, USA}

\sysmlcorrespondingauthor{Sheng-Chun Kao}{felix@gatech.edu}

\sysmlkeywords{Neural Architecture Search, GAN, Quantization, Generative model}

\vskip 0.3in

\begin{abstract}
Designing resource-efficient Deep Neural Networks (DNNs) is critical to deploy deep learning solutions over edge platforms due to diverse performance, power, and memory budgets.
Unfortunately, it is often the case a well-trained ML model does not fit to the constraint of deploying edge platforms, causing a long iteration of model reduction and retraining process. 
Moreover, a ML model optimized for platform-A often may not be suitable when we deploy it on another platform-B, causing another iteration of model retraining. 


We propose a conditional neural architecture search method using GAN, which produces feasible ML models for different platforms.
We present a new workflow to generate constraint-optimized DNN models. 
This is the first work of bringing in condition and adversarial technique into Neural Architecture Search domain. We verify the method with regression problems and classification on CIFAR-10. The proposed workflow can successfully generate resource-optimized MLP or CNN-based networks.
\end{abstract}

]



\printAffiliationsAndNotice{}  

\section{Introduction}
Deep Neural Networks (DNN) have seen tremendous success in solving several problems such as regression, image classifications, object detection, sequential decision making, and so on. Owing to this, and the growth in personal and cloud computing capabilities, machine learning (ML) practitioners 
today design deep learning architectures with millions of parameters that require billions of computations. 
With the advent of edge computing, there has been a significant thrust towards the deployment of such complex DNNs on devices with limited computing power such as mobiles and embedded systems to perform real-time data analytic on perceived and sensed data, as in the case of object detection in autonomous cars. Realization of such deployments, though, relies on extremely high resource- and energy-efficiencies, and has led to large number of the recent works addressing the issue of hardware deployment by model compression or quantization \cite{he2018amc, yang2018netadapt, liu2017learning, han2015deep, zhou2017incremental}.

However these model compression technique often relies on a pre-defined DNN such as MobilenetV2 ~\cite{sandler2018mobilenetv2}, which limit the capability of finding other DNN model structures that may reach better performance within the same constraint. This is a common dilemma in practice since ML designer and hardware practitioners are often in different entities in an organization. There is little, if any, product tuning iteration in the process of model deployment. When iterations do occur, they result in tedious retraining and re-deployments that are extremely time consuming.

\begin{figure*}[ht]
\begin{center}
\includegraphics[width=1.0\linewidth]{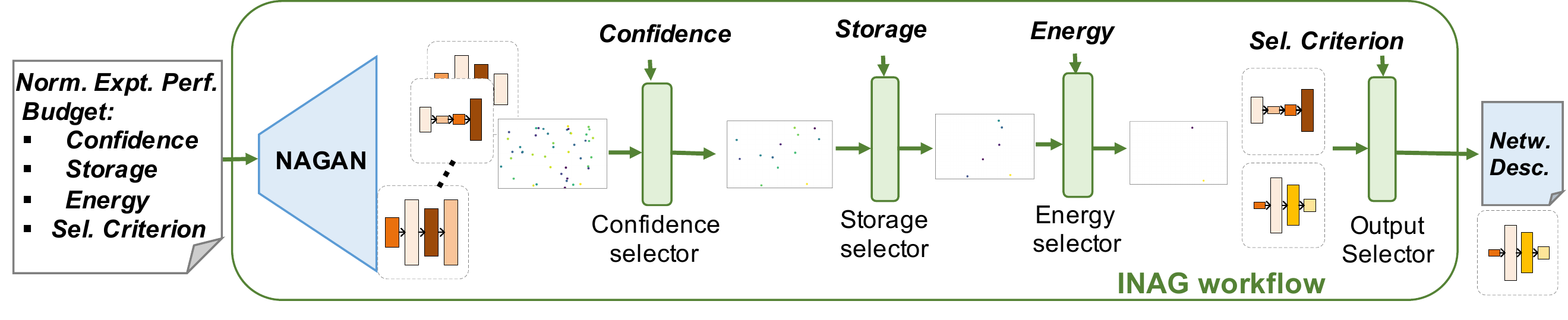}
\end{center}
\vspace{-0.40cm}
   \caption{The workflow of Inverse Neural Architecture Generation (INAG). A set of quantized neural architectures that fit the model accuracy target are generated by Neural Architecture GAN (NAGAN) and piped to multiple selecting stages for different resource constraints.}
\vspace{-0.2cm}
\label{fig:workflow}
\end{figure*}

Therefore many Platform-aware Neural Architecture Search (NAS) ~\cite{linneural, mnasnet} techniques have been proposed, which search the neural architecture that fits a certain platform. However, the found ML model is often optimized for a specific platform. A new search procedure is needed when the underlying platform changes, which implies the found model is not portable from platform to platform.

Inspired by the wide success in Conditional Generative Adversarial Networks (cGAN), we introduce Neural Architecture GAN (NAGAN), which when conditioned on the expected model performance\footnote{Model performance in this paper refers to accuracy for classification problems, and R-square error for regression problems (details in \autoref{sec:exp_setup}).}generates multiple feasible network architectures with their quantization bit-width. NAGAN generates various structure of NNs which shares similar performance. Therefore, when switching the platform, we can choose a different NN that fits the constraint from the generated NNs pool. Thus, NAGAN can generate NNs for unknown platforms, which is essential in the era when the new platforms are coming out every day. Moreover, NAGAN adopts to both high-end and low-end platform. While performance is expected to be traded-off by area and power, NAGAN generate lighter NNs by conditioning on lower expected model performance.

We also present an Inverse Neural Architecture Generation (INAG) workflow, which demonstrates a practical design flow when applying conditional Neural Architecture Search method to the resource constraint problem, as shown in \autoref{fig:workflow}.

\begin{table}[ht]
\begin{center}
  \caption{The comparisons of state-of-the-art works}
\includegraphics[width=1\linewidth]{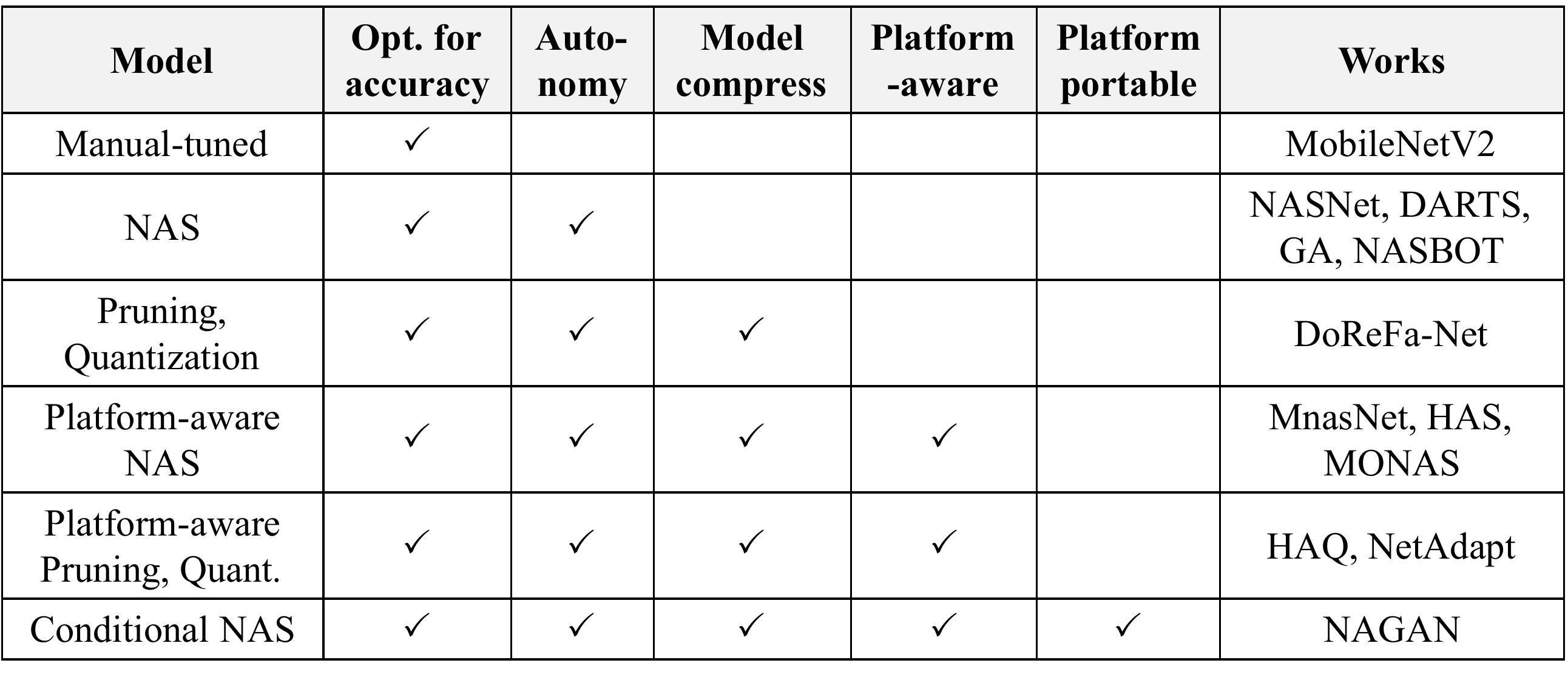}
\end{center}
\vspace{-0.40cm}

\vspace{-0.2cm}
\label{table:compare_table}
\end{table}

To verify the proposed method, we apply NAGAN to Multi-Level Perceptron (MLP) and Convolution-based Neural Network (CNN) architecture search on regression problems and classification problems on both synthetic dataset, and real-worlds dataset, including MNIST and CIFAR-10. The NAGAN can generate a bag of NNs within 10\% difference to the conditioning value. Then we show how to apply INAG workflow to choose feasible NNs. We recognized that there are many unexplored challenges in this framework such as extending to larger dataset, supporting more diverse NNs sturctures, and scalability to more complex problem. We intend to further investigate these topics in our future work. However, this work is a preliminary step to incorporate condition and adversarial technique to the NAS problem, which demonstrates a potential path toward the variety of NAS applications.

 
The primary contributions of this work are as follows:
\squishlist
    \item This is the first work, to the best of our knowledge, to propose a conditional GAN-based neural architecture search. 
    
    \item Our conditional neural architecture search framework optimizes the network architecture and layer-wise quantization bit-width simultaneously.
    
    
    \item We present a practical workflow to utilized the conditional neural architecture search method.
    
    \item Our end-to-end workflow learns the mapping between performance and neural architectures and can inversely generate the architecture for the desired performance.

\squishend




\section{Background and Motivation}
\label{sec:background}

Resource constraints on edge-devices have led to active interest in designing DNN models which trade-off their accuracy for lower compute and memory footprint~\cite{squeezenet, sandler2018mobilenetv2, zhang2018shufflenet}. We describe related work in \autoref{sec:related}. 
We also qualitatively compare the contributions of relevant state-of-the-art works in \autoref{table:compare_table}.

\subsection{Training Quantizated Networks}
Quantized DNN models quantize weights and activations to represent the floating number in lower bit-widths \cite{courbariaux2015binaryconnect, rastegari2016xnor, zhou2017incremental}.
%
In this work, we target edge platforms where floating point arithmetic is expensive. We follow the concept of integer-arithmetic-only quantization \cite{int_Quantize}, which presents all parameters arithmetic in integers. We leverage this trainable quantization method in this work to learn the  network architecture and quantization bit-width simultaneously.


\subsection{Multi-objective NAS}
Neural Architecture Search (NAS) automates the search process of the model. 
Most NAS works can be formulated as a hyper-parameter search problem. While most NAS works are optimizing the accuracy, Platform-aware NAS \cite{mnasnet, tan2019efficientnet, kim2017nemo} jointly optimize accuracy and platform constraint by adding the platform constraint into the target function, called multi-objective optimization.

However, Multi-objective NAS faced two challenges: (1) the objective function need careful manual design to represent different platform constraint; (2) a found platform-optimized NN may not be portable to other platform. 

\subsection{Generative Adversarial Networks (GAN)}
Generative Adversarial Nets (GANs) \cite{goodfellow2014generative}
contains a generative network (\textit{generator}) and an adversarial network (\textit{discriminator}). The generator is pitted against the discriminator, which learns to determine whether a sample is from the generator or the training data. Generator and discriminator are trained iteratively to outperform each other, thus resulting in both models improving over the course of the training process. 


Conditional GAN \cite{mirza2014conditional, van2016conditional} takes in the condition in the training phase, so that the mapping of different condition to different distribution can be learned. Similar approaches such as ACGAN \cite{acgan}, infoGAN \cite{infogan}, and others \cite{odena2016semi, ramsundar2015massively}, task the discriminator to reconstruct the condition taken by the generator.





\section{Neural Architecture GAN}
\vspace{-2mm}
This section illustrates the architecture of the proposed Neural Architecture GAN (NAGAN), which is the first stage of the proposed INAG workflow, described in \autoref{sec:inag}.

\subsection{NAGAN Overview}
\label{sec:nagan_overview}
\begin{figure}[!t]
\begin{center}
\includegraphics[width=1.0\linewidth]{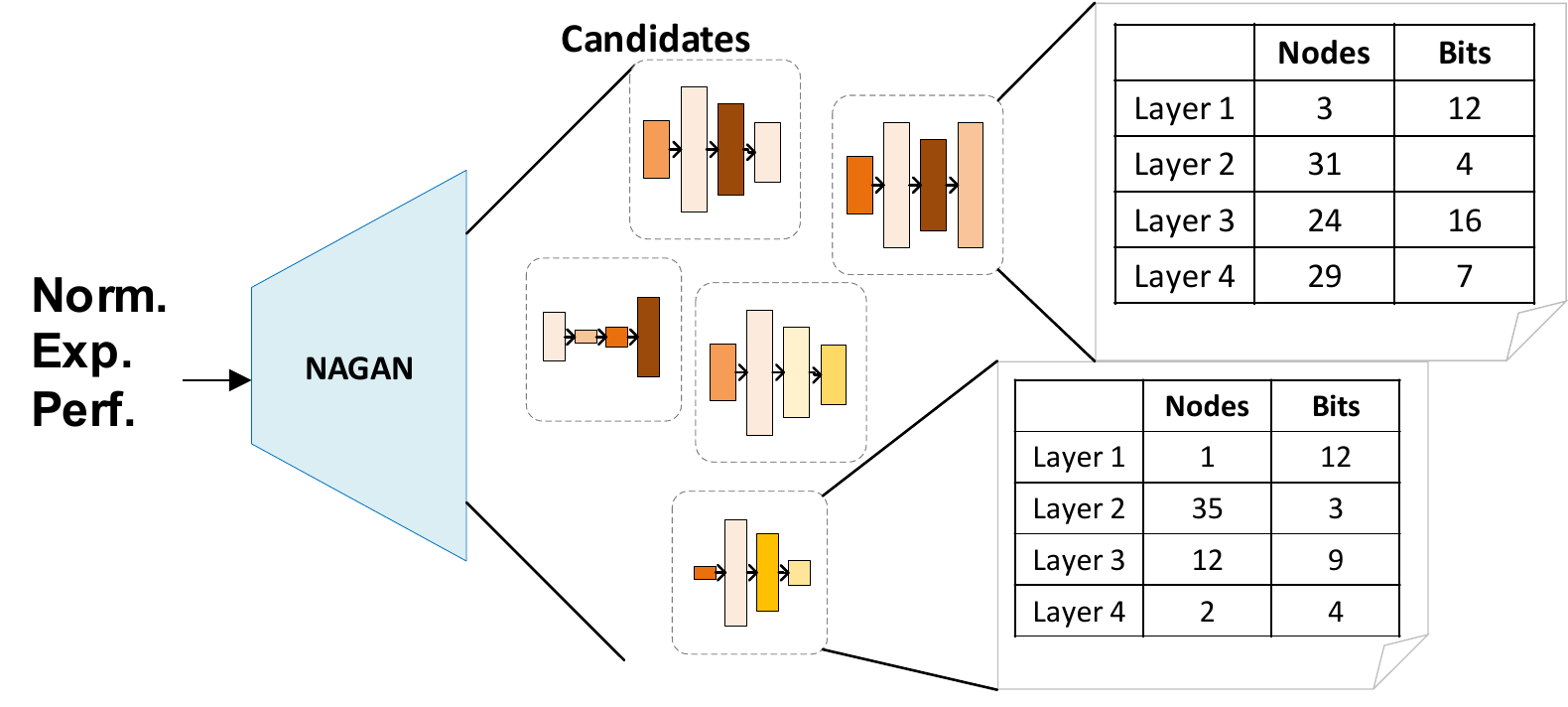}
\end{center}
\vspace{-0.40cm}
   \caption{The generator of NAGAN takes in a condition (normalized expected performance) and generate a set of feasible neural architectures. }
\vspace{-0.1cm}
\label{fig:generator}
\end{figure}

NAGAN, 
 takes in the normalized expected model performance as its input, and generates a set of feasible neural architectures, each with a per-layer quantization, as shown in \autoref{fig:generator}. The training of NAGAN contains a generator, a discriminator, and an encoder. The generator and discriminator are inherited from the general framework of conditional GAN, involving modifications that enable them to learn the distribution of neural architectures.

\begin{figure}[!t]
\begin{center}
\includegraphics[width=1.0\linewidth]{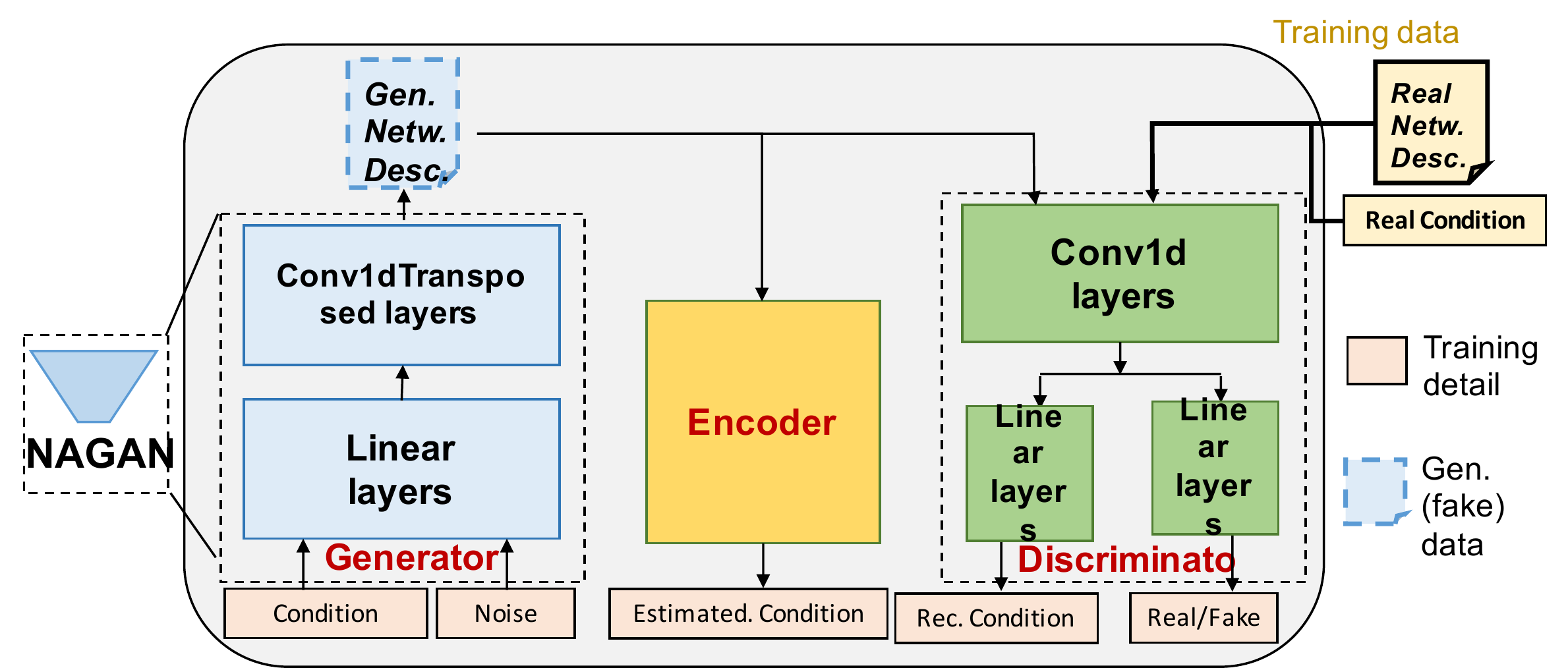}
\end{center}
\vspace{-0.40cm}
   \caption{NAGAN Training.}
\vspace{-0.2cm}
\label{fig:nagan}
\end{figure}

 \textbf{Encoder.} As shown in \autoref{fig:nagan}, we add an encoder to enhance the training performance of generator. The encoder is trained by the model definition to model performance pair with regression manner. The main purpose of the encoder is the fast estimation of the quality of the generated data. While quality evaluation in the image generation tasks of GANs are not standardized and mostly rely on human judgement, we found that the quality of the network architecture can be systematically judged by applying the generated network descriptions to the platforms and collecting their performance. However, as the addressed challenge of platform-aware NAS \cite{mnasnet}, to interact with the real platform can be time-consuming, an appropriate surrogate model that predicts the performance is needed for fast interaction. NAGAN trains an encoder to serve as a surrogate model.

\vspace{-1mm}

The encoder is a pre-trained network that maps network description to the network performance. The encoder is found to be helpful in several ways in the training process. First, before actually training NAGAN, training the encoder provides a means for the fast estimation of how the NAGAN may perform, e.g., we can notice early if the number of training data is inadequate. Second, the pre-trained encoder weights can be a good initiation for the first few layers of discriminator since the encoder and discriminator work on the same feature map. 


\subsection{Training of NAGAN}
\label{sec:nagan_training}
\vspace{-1mm}
The inputs of the generator are a Gaussian noise on latent dimensions and a fake (generated) condition. For example, assuming the dimensions of latent space is 10, and the dimension of condition is 1, whose value ranges from 0 to 1 after normalization, then in the training process, we randomly sample a 10 dimension Gaussian noise and a value from 0 to 1 as inputs of latent space and condition 
respectively for the generator as shown in bottom left of \autoref{fig:nagan}. The discriminator has two source of inputs, the generated data (generated network description) and the training data (a set of real network description with its corresponding normalized performance value). The discriminator is trained to accomplish two tasks: first, to tell whether the input data is from generated set or training set and second, to reconstruct the condition of the input data. As shown in Algorithm \autoref{alg:training}, we alternate between the training of the generator, discriminator, and encoder by optimizing their value functions.


\textbf{Value functions.} 
The value function of generator and discriminator conform to the ACGAN \cite{acgan} setting. On top of it, we add another term of encoder's value function $V(E)$ in the training process. In our framework, the weights of the encoder are pre-trained and fixed. $V(E)$ evaluates the loss of generated NNs' condition approximated by the surrogate model, encoder. The loss of $V(E)$ will back-propagates through and trains the generator. Encoder is a teaching model for generator.

\begin{algorithm}[tb]
   \caption{NAGAN Training}
   \label{alg:training}
\begin{algorithmic}
   \STATE {\bfseries Input:} I: number of iterations
   \FOR{$i=1$ {\bfseries to} $I-1$}
   \STATE \textbf{Sample} $m$ positive example \\$\left \{(c^{0}, x^{0}), (c^{1}, x^{1}),..., (c^{m-1}, x^{m-1})\right \}$
  \STATE  \textbf{Sample} $m$ noise sample $\left \{z^{0}, z^{1},..., z^{m-1}\right \}$, generate data $\left \{\tilde{x}^{0}, \tilde{x}^{2},..., \tilde{x}^{m-1}\right \}, \tilde{x}^{i}=G(c^{i}, z^{i})$ and reconstruct labels $\left \{\bar{c}^{0},\bar{c}^{1}...\bar{c}^{m-1}\right \}$ and $\left \{\tilde{c}^{0},\tilde{c}^{1}...\tilde{c}^{m-1}\right \}$, $\bar{c}^{i}=D_{2}(x^{})$, $\tilde{c}^{i}=D_{2}(\tilde{x}^{i})$
   \STATE \textbf{Update} discriminator V(D) via gradient ascent
   \STATE \textbf{Sample} $m$ noise sample, generated data and reconstruct labels 
   \STATE \textbf{Update} generator V(G) via gradient descent
   \STATE \textbf{Sample} $m$ noise sample, generated data and reconstruct labels 
   \STATE \textbf{Update} encoder V(E) via gradient descent
   \ENDFOR
\end{algorithmic}
\end{algorithm}

\textbf{Continuous condition.}
We use the model performance as a continuous condition.
We found that it encourages a smooth transition along the dimension of condition. The effectiveness of continuous condition is also seen in prior work\cite{diamant2019beholder, pumarola2018ganimation, souza2018gan}.




\textbf{Training Data.} Since there is no public training data available for the task of neural architecture generation, we prepare the training data as follows. Each training data for NAGAN is composed of a network description (number of nodes/channels and quantization bit-width per layer) and its corresponding model performance. We randomly sample the value for each parameter in the network description, deploy and train it on the regression or classification problem, and gather its model performance e.g., accuracy. We execute a Monte Carlo sampling of the network description space to generate sufficient data to compose the training dataset.

\subsection{Generating Quantized Neural Architectures}
To achieve best data efficiency for the neural architecture, our goal is to not only produce the network architecture but also the quantized bit-width for each layer of the generated networks. There is existing research that targets uniform quantization of the whole model (model-wise quantization) and reports impressive result on 8-bit quantization for tested applications \cite{gysel2018ristretto}. In this work, we target per-layer quantization, as shown in the example in \autoref{fig:generator}, which reduce the model size more aggressively than uniform quantization.

\section{Inverse Neural Arch Generation}
\label{sec:inag}
\vspace{-1mm}

The NAGAN generates a bag of NNs that fit the required condition. Out of these NNs, we pick the one that fits the platform. In this section, we present a practical workflow called Inverse Neural Architecture Generation (INAG) to accomplish the selection process.


\vspace{-2mm}
\subsection{Tool-flow and Walk-through example of INAG.}

After NAGAN generated a bag of NNs, the INAG simply select these NNs by different criterion of setting platform constraint. However, when no generated NNs fits the constraint (for example, when we require to put a high complexity NN on small edge device), we condition on lower expected model performance and generate a new bag of lower-end NNs to fit in lower-end platform.



\subsection{Selecting Stages}
Selecting stages are fast evaluation processes, which output the data points meeting the specified set of constraints. Unlike the state-of-the-art multi-objective optimization approaches, we consider the constraint \textit{after} the candidate feasible data points are generated by NAGAN. Moreover, selecting stages remove the necessity for designing weighting parameters for each constraint in the multi-objective problem which leads to distinct optimum points automatically when the constraint changes. 

\textbf{Confidence selector.}
\label{subsec:confidence}
Since NAGAN generates the models from Gaussian sampling, the wider accepted confidence range leads to higher diversities of models and likewise larger range of performance difference in a bag of NNs. However, we can evaluate the confidence value of each NN by calculating the normalized distance, $Dist_{f}(m)$, between the input condition $c$ and the evaluated condition of generated NN, $m$. INAG uses the encoder as a surrogate model to estimate the performance, $Test_{f}(m)$. The normalized distance is defined as:
\begin{flalign}
\hskip\parindent & \begin{gathered}
    Dist_{f}(m)=\frac{1}{P_{R}}\left | Test_{f}(m)-c \right |,
    \end{gathered} &
\end{flalign}
where $P_{R}$ is the normalized factor to normalize $c$ between 0 and 1. The confidence selectors use the normalized distance as its confidence for selecting the feasible NNs.


\textbf{Storage selector.}
The storage requirement to deploy a DNN on a device is formulated as a function of number of parameters and their bit-widths, which approximates the required memory space. The evaluation function is formulated as:
\begin{flalign}
\hskip\parindent & \begin{gathered}
    Storage(m) = \sum_{i=0}^{N-1}    \left [ W(i) + F(i) \right ]B(i),
    \end{gathered} &
\end{flalign}
where $N$ is the number of layers of model $m$. $W(i)$ is the number of weight parameters, $F(i)$ is the number of feature map parameters, and $B(i)$ is the bit-width of layer $i$. The INAG workflow presents normalized storage constraint, where the value is normalized to the range of 0 to 1.
 
\textbf{Energy selector.}
 Since the MACs dominates the operating energy \cite{yang2017designing}, and the power of one MAC unit is a function of arithmetic bit-width, INAG 
formulates the operating energy consumption as a function of MACs and bit-width:
\begin{flalign}
\hskip\parindent & \begin{gathered}
    Energy(m)=\sum_{i=0}^{N-1} f(MAC(i), BIT(i)).
    \end{gathered} &
\end{flalign}
$MAC(i)$ is the total number of MAC operations of layer $i$. The $f(x, y)$ is the energy evaluate function. Also we used normalized energy value between 0 and 1 in the whole flow.

\textbf{Output selector.} 
The selector is the last-stage of INAG, which selects the highest ranked generated design configuration, based on the criterion of  $Dist_{f}(m)$, normalized storage, or energy.

\subsection{Extensions}
In practice, INAG can easily incorporate additional constraints such as end-to-end performance, main-memory bandwidth, and energy 
on a target device 
by enhancing or adding 
more selecting stages that leverage advanced external
analytical tools for estimating latency or throughput~\cite{maestro}, optimal dataflow~\cite{maestro}, energy~\cite{accelergy}, and so on.

\section{Evaluation}
\subsection{Experiment setup}
\label{sec:exp_setup}
\textbf{Dataset.} 
For the regression problems, we use two types of datasets: synthetic and real world. For the synthetic datasets, we generate two datasets with different complexity of the synthetic functions $f$ with additive Gaussian noise to the function, which is a common assumption for most regression models, where the synthetic function for Dataset $i$ becomes $Data_{i}(x) = f_{i}(x) + z$. We have $z$ as an Gaussian noise. For the first dataset $Data_{a}$, we consider a polynomial function of degree of $4$ with randomly assigned coefficients. In particular, we have $f_{1}(x)=-2x^{4}-8x^{3}+5x^{2}+15$ for $Data_{a}$ in the presented result. For the second dataset $Data_{b}$, we consider a more complex polynomial function of degree of $7$.  In particular, we have $f_{2}(x)=-2x^{7}+2x^{5}-4x^{3}+15$ for $Data_{b}$. We synthesize these two datasets to showcase that our method is effective in both simple and complex regression problems. For the real word dataset, we apply California Housing~\cite{CH}, whose task is to predict the price of a property given the property attribute. For the image classification problems, we use two datasets: MNIST and CIFAR-10.

\textbf{Target Problems.}
We construct six different setting of experiments for Regression and Classification with MLP-based and CNN-based structure and named them accordingly in \autoref{table:desc_table}.

\begin{table}[]
\scriptsize
\caption{The Experiment settings.}
\begin{tabular}{|c|c|c|c|c|c|c|}
\rowcolor[HTML]{EFEFEF} 
\textbf{Exp.} & \textbf{Reg.A} & \textbf{Reg.B} & \textbf{Reg. C} & \textbf{Cls.A} & \textbf{Cls.B} & \textbf{Cls.C} \\
\cellcolor[HTML]{EFEFEF}\textbf{Arch.} & MLP & MLP & MLP & MLP & MLP & CNN \\
\cellcolor[HTML]{EFEFEF}\textbf{Data} & $Data_{a}$ & $Data_{b}$ & C. H. & MNIST & CIFAR & CIFAR
\end{tabular}
\label{table:desc_table}
\end{table}

\textbf{Evaluation metrics.} 

When evaluating the performance of the NAGAN generator, we statistically calculate the normalized distance of the generated model $Dist(m)$ and the normalized condition. We sweep the condition from 0.1 to 1.0 using a step size of 0.1. The generator's performance is evaluated by the mean normalized distance, $mDist_(G)=\frac{1}{10P_{R}}\sum_{i=1}^{10}Test(G(P_{t}(i)))-P_{t}(i)$. In the experiment, we present two testing platforms. The first is the fast evaluation metric by the encoder $Test_{f}(G)$ as shown in \autoref{subsec:confidence}, which gives $mDist_{f}(G)$. We also evaluate the actual performance $Test_{r}(G)$ by applying the generated NNs to the platform, which gives $mDist_{r}(G)$.    

\begin{figure}  
\begin{center}
\includegraphics[width=1\linewidth]{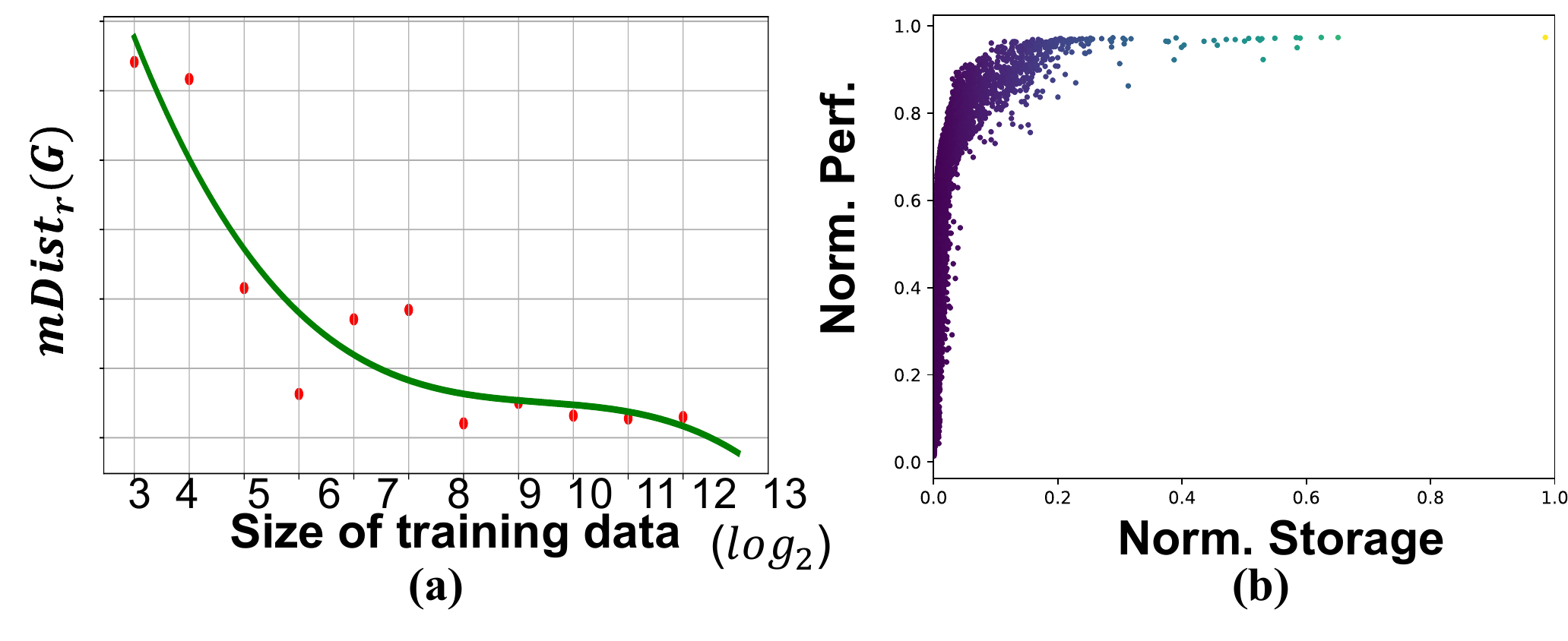}
\end{center}
\vspace{-0.40cm}
\caption{In Exp. Reg.A, (a) the $mDist_{r}(G)$ when trained with different number of data \textsuperscript{*}, and (b) the scatter plot showing Pareto frontier on normalized performance and storage.}
\label{fig:fig4}
\scriptsize \textsuperscript{*} Note: Red dot is the actual value. The green line is a smoothed trend curve.
\vspace{-0.10cm}
\end{figure}

\textbf{Size of Training Data.} The neural architecture search space is extremely large. For instance, in our setting, the total number of datapoints in the search space for the MLP architecture generation is more than 17 billions.
However, empirically we found that we can train NAGAN with a comparatively small training set. In \autoref{fig:fig4}(a), we show the impact of the number of training data samples to the quality of NAGAN via the example of Reg.A. We trained NAGAN with different size of training data and measured the $mDist_{r}(G)$. We found that $mDist_{r}(G)$ starts to converge around 1k datapoints. This implies we can train NAGAN with small number of data and achieve good performance close to the one trained with large number of data. In the following experiments, we use training set size as follows: 8,000 datapoints in $Data_{a}$, 8,000 in $Data_{b}$, 7,000 in California Housing, 10,000 in MNIST, and 20,000 in CIFAR-10.


\begin{table}[]
\vspace{-0.5cm}
\caption{The comparison results on Reg.A dataset.}
\begin{tabular}{|c|c|c|c|c|}
\hline
\rowcolor[HTML]{EFEFEF} 
 & Perf.(\%) & Storage & Energy &  Time  \\ \hline
\cellcolor[HTML]{EFEFEF}\textbf{GA} & 97.3 & 0.21 & 0.21 & 6 hrs  \\ \hline
\cellcolor[HTML]{EFEFEF}\textbf{Bayesian} & 97.7 & 0.20 & 0.213 & 13 hrs \\ \hline
\cellcolor[HTML]{EFEFEF}\textbf{INAG} & 95 & 0.19 & 0.19 & 5 secs \\ \hline

\end{tabular}
\scriptsize \textsuperscript{*} Note: The constraint in this experiment is defined as: Normalized storage \textless{} 0.21, Normalized Energy \textless{} 0.21. Time is the searched time for GA and Bayesian, and the inference time of INAG to get this result while a pre-trained NAGAN is trained in around 2 hrs.
\label{table:exp_table}
\end{table}



\subsection{Comparison with GA and Bayesian Optimization}
\label{subsec:comparison}
The results of the INAG workflow are compared against the baseline approaches of genetic algorithm (GA) and Bayesian optimization. 
We perform a constrained optimization using a GA in which the algorithm attempts to maximize the accuracy of the model subject to constraints of storage and energy. The constraint is incorporated into the fitness of each design point using the interior penalty method. 
We similarly use Bayesian optimizer to search the optimum point by incorporating the constraint into target function. The search time of each method is measured with a desktop with a GTX1080 GPU, as shown in \autoref{table:exp_table}.


Further, a trained NAGAN is portable between platform. We can use NAGAN to re-generate NNs and go through the fast selection of INAG workflow whenever the platform changes. However the optimization-based manner such as GA and Bayesian need redesigning of the target function and go through the search process again.     

\subsection{Results on NAGAN}
 we also statistically evaluate NAGAN by the defined evaluation metric  $mDist_{f}(G)$ and  $mDist_{r}(G)$ as illustrated in \autoref{sec:exp_setup}. Across all experiments in \autoref{table:desc_table}, the average of actual mean normalized distance of generator $mDist_{r}(G)$ is below 10 \%, as shown in in \autoref{sec:exp_setup}. We can observe that $mDist_{f}$ is smaller than $mDist_{r}$, which means the surrogate model, encoder, is sometime overly optimistic. However, it stills helps as a fast evaluation in the NAGAN training process and INAG selecting process and helps navigate the NAGAN through the training.


\textbf{Discussion.} From the experiment Cls.C, we found that the generator was capable of capturing the distribution of the CNN, yielding comparable accuracies to its MLP counterpart. Even though both the search space of the CNN is larger and the latent distribution being more complex, we observe that the NAGAN is able to achieve a good $mDist_{r}(G)$ of 8.7\% on the CIFAR-10 dataset. Although we demonstrate the capability of NAGAN to capture the complexities in CNNs, we recognize that the demonstrated results shown in \autoref{table:error} is more data intensive requiring 20,000 training datapoints to reach the quoted performance. Hence, there is still plenty of room for exploration, e.g., exploiting of different characteristics of CNN, supporting more complex CNN models. We intend to further investigate these topics in our future work.

\begin{table}[]
\caption{The evaluation of NAGAN.}
\begin{small}
\begin{tabular}{ccccccc}
\hline

\rowcolor[HTML]{EFEFEF} 
\multicolumn{1}{|c|}{\cellcolor[HTML]{EFEFEF}Exp} & \multicolumn{1}{c|}{\cellcolor[HTML]{EFEFEF}Reg.A} & \multicolumn{1}{c|}{\cellcolor[HTML]{EFEFEF}Reg.B} & \multicolumn{1}{c|}{\cellcolor[HTML]{EFEFEF}Reg.C} & \multicolumn{1}{c|}{\cellcolor[HTML]{EFEFEF}Cls.A} & \multicolumn{1}{c|}{\cellcolor[HTML]{EFEFEF}Cls.B}& \multicolumn{1}{c|}{\cellcolor[HTML]{EFEFEF}Cls.C} \\ \hline
\multicolumn{1}{|c|}{\cellcolor[HTML]{EFEFEF}$mDist_{f}$} & \multicolumn{1}{c|}{0.3} & \multicolumn{1}{c|}{1.2} & \multicolumn{1}{c|}{0.5} & \multicolumn{1}{c|}{0.3} & \multicolumn{1}{c|}{0.4} & \multicolumn{1}{c|}{4.8} \\ \hline
\multicolumn{1}{|c|}{\cellcolor[HTML]{EFEFEF}$mDis_{r}$} & \multicolumn{1}{c|}{12.0} & \multicolumn{1}{c|}{6.2} & \multicolumn{1}{c|}{11.3} & \multicolumn{1}{c|}{8.1} & \multicolumn{1}{c|}{11.7} & \multicolumn{1}{c|}{8.7} \\ \hline
\multicolumn{7}{l}{}
\end{tabular}
\end{small}
\label{table:error}
\end{table}

\subsection{Results on INAG}
\label{subsec:result_inag}
We demonstrate the example of INAG by Reg.A to show its advantage of wide Pareto frontier. \autoref{fig:fig4}(b) shows a scatter plot of normalized performance to storage when we sweep the expected normalize performance from 0.0 to 1.0. It shows that INAG can actually generate different NNs with different storage consumption conditioning on its expected normalize performance. Also the generated NNs forms a wide Pareto frontier. 
\section{Related Works}
\label{sec:related}
\vspace{-1mm}


\textbf{Quantization.}
Deep Compression \cite{han2015deep} quantized the values into bins, where each bin shares the same weight, and only a small number of indices are required. Courbariaux et al., \cite{courbariaux2014training} directly shifted the floating point to fixed point and integer value. 
DoReFa-Net \cite{zhou2016dorefa} retrained the network after quantization and enabled the quantized backward propagation. 
HAQ \cite{wang2019haq} approached the quantized network training with reinforcement learning.


\textbf{Neural Architecture Search (NAS).}
The genetic algorithms (GA) and evolution algorithms \cite{stanley2019designing} have been studied for decades for parameter optimization. NASBOT \cite{kandasamy2018neural} applied Bayesian optimisation on architecture search. NAS with reinforcement learning are the most general framework currently such as NASNet \cite{zoph2018learning} . DARTS \cite{darts} used differential architecture to facilitate the search process. Platform-aware NAS, such as MnasNet \cite{mnasnet}, MONAS \cite{hsu2018monas}, HAS \cite{linneural}, and NetAdapt \cite{yang2018netadapt} included platform constraint in the search process. 
Unlike the previous efforts of NAS, this work utilized the conditional method to jointly generate the network architecture and quantized bit-width conditioning on different expected performance for different platforms.

\section{Conclusion}
\vspace{-2mm}
We propose a conditional-based NAS method to provide platform portability. Also, we present a workflow to select the feasible NNs for the targeted platform. Finally, we validate our method on both regression and classification problem. 


\nocite{langley00}

\bibliography{main}
\bibliographystyle{sysml2019}



\end{document}